\documentclass{article}
\usepackage{amssymb}

\usepackage[final]{corl_2025} %

\usepackage{graphicx}          %
\usepackage[table,dvipsnames]{xcolor}  %
\usepackage{url}               %
\usepackage{booktabs}          %
\usepackage{microtype}         %

\usepackage{amsmath, amssymb, amsfonts, amsthm}  %
\usepackage{nicefrac}          %
\usepackage{bm}                %
\usepackage{bbm}               %

\usepackage{tabularx}          %
\usepackage{multirow}          %
\usepackage{floatrow}          %
\usepackage{colortbl}          %

\usepackage[ruled, vlined, linesnumbered]{algorithm2e}  %
\DontPrintSemicolon  %

\usepackage{subcaption}        %
\usepackage{wrapfig}           %
\usepackage{adjustbox}         %

\usepackage[titletoc]{appendix}
\usepackage{titletoc}
\usepackage{etoc}              %
\usepackage{multicol}          %
\usepackage{chngpage}          %
\usepackage{soul}              %
\usepackage{csquotes}          %
\usepackage{yfonts}            %

\usepackage{mdframed} %
\usepackage{listings}%
\definecolor{light-gray}{gray}{0.95}
\usepackage{thmtools, thm-restate}  %

\usepackage{tikz}              %

\usepackage{hyperref}          %

\usepackage[numbers]{natbib}   %

\usepackage[capitalise,nameinlink]{cleveref}

\definecolor{flodarkpurple}{rgb}{0.288,0.1196,0.7}
\definecolor{RedOrange}{rgb}{0.9, 0.3, 0.0}

\newcommand{\bcl}[1]{\cellcolor{blue!13}#1}

\newcommand{\btext}[1]{\textcolor{blue!60}{#1}}

\newcommand{\encoder}{Enc}
\newcommand{\decoder}{Dec}
\newcommand{\dynamics}{Dyn}
\newcommand{\dataset}{\mathcal{D}}

\newcommand{\SkillSeq}{\phi_{1:H}(a_{1:H})}
\newcommand{\render}{\Psi}

\newcommand{\rd}{\Gamma}
\newcommand{\stein}{SVI}
\newcommand{\svgd}{Fail2Progress}
\newcommand{\gradient}{Gradient}
\newcommand{\sampling}{Sampling}
\newcommand{\replan}{Replanning}
\newcommand{\orig}{Original}
\newcommand{\sma}{Small}
\newcommand{\lar}{Large}
\newcommand{\point}{Points2Plans}

\newcommand{\fail}{F}

\usepackage{subcaption}
\usepackage[skip=0pt,font=small,labelfont=bf]{caption}
\usepackage{amsmath}
\DeclareMathOperator*{\argmax}{arg\,max}
\DeclareMathOperator*{\argmin}{arg\,min}

\long\def\meandist#1#2{\mathbb{E}_{#2}\left[#1\right]}
\long\def\dkl#1#2{D_{KL}\Big( #1 \; \big\| \; #2\Big)}

\definecolor{revision_blue}{RGB}{0, 0, 200}

\usepackage[subtle]{savetrees} %
\title{Fail2Progress: Learning from Real-World Robot Failures with Stein Variational Inference}

\definecolor{flodarkpurple}{rgb}{0.288,0.1196,0.7}
\newcommand{\authorhref}[3][flodarkpurple]{\href{#2}{\textcolor{#1}{#3}}}
\author{%
    \normalfont
    \authorhref{https://yixuanhuang98.github.io/}{Yixuan Huang}\textsuperscript{1},\,
    \authorhref{https://novellaalvina.github.io/}{Novella Alvina}\textsuperscript{1},\,
    \authorhref{https://users.cs.utah.edu/~mohanraj/}{Mohanraj Devendran Shanthi}\textsuperscript{1},\,
    \authorhref{https://robot-learning.cs.utah.edu/thermans}{Tucker Hermans}\textsuperscript{1,2}\\\vspace{-6pt}\\
    \textsuperscript{1}\href{https://www.utah.edu/}{University of Utah},
    \textsuperscript{2}\href{https://www.nvidia.com/en-us/research/}{NVIDIA Research}
    \vspace{-20pt}
}

\begin{document}
\maketitle

\begin{abstract}
Skill effect models for long-horizon manipulation tasks are prone to failures in conditions not covered by training data distributions. 
Therefore, enabling robots to reason about and learn from failures is necessary.  
We investigate the problem of efficiently generating a dataset targeted to observed failures. 
After fine-tuning a skill effect model on this dataset, we evaluate the extent to which the model can recover from failures and minimize future failures. 
We propose \svgd{}, an approach that leverages Stein variational inference to generate multiple simulation environments in parallel, enabling efficient data sample generation similar to observed failures. 
Our method is capable of handling several challenging mobile manipulation tasks, including transporting multiple objects, organizing a constrained shelf, and tabletop organization.
Through large-scale simulation and real-world experiments, we demonstrate that our approach excels at learning from failures across different numbers of objects. 
Furthermore, we show that \svgd{} outperforms several baselines. 
Qualitative results are available at \href{https://sites.google.com/view/fail2progress}{sites.google.com/view/fail2progress}. 

\end{abstract}

\keywords{Learning from failures, Variational inference, Skill effect models} 

\section{Introduction}
Learned models of skill effects~\cite{liang-icra2022, huang2024points2plans, chen2023predicting, paxton-corl2021-semantic-placement, stap} show promising results in solving long-horizon manipulation tasks via skill sequencing. 
To train these models, researchers typically leverage simulation to efficiently generate large-scale, diverse data. 
However, robots using skill-based models in unstructured and uncertain real-world environments will inevitably struggle in out-of-distribution scenarios that are significantly different from the training datasets. 
In response, we want our robots to detect failures, recover from failures, and learn to minimize future failures so that they can continuously adapt once deployed. 

\begin{figure}[ht]
    \centering
   \includegraphics[width=0.99\columnwidth,clip,trim=0mm 0mm 0mm 0mm]{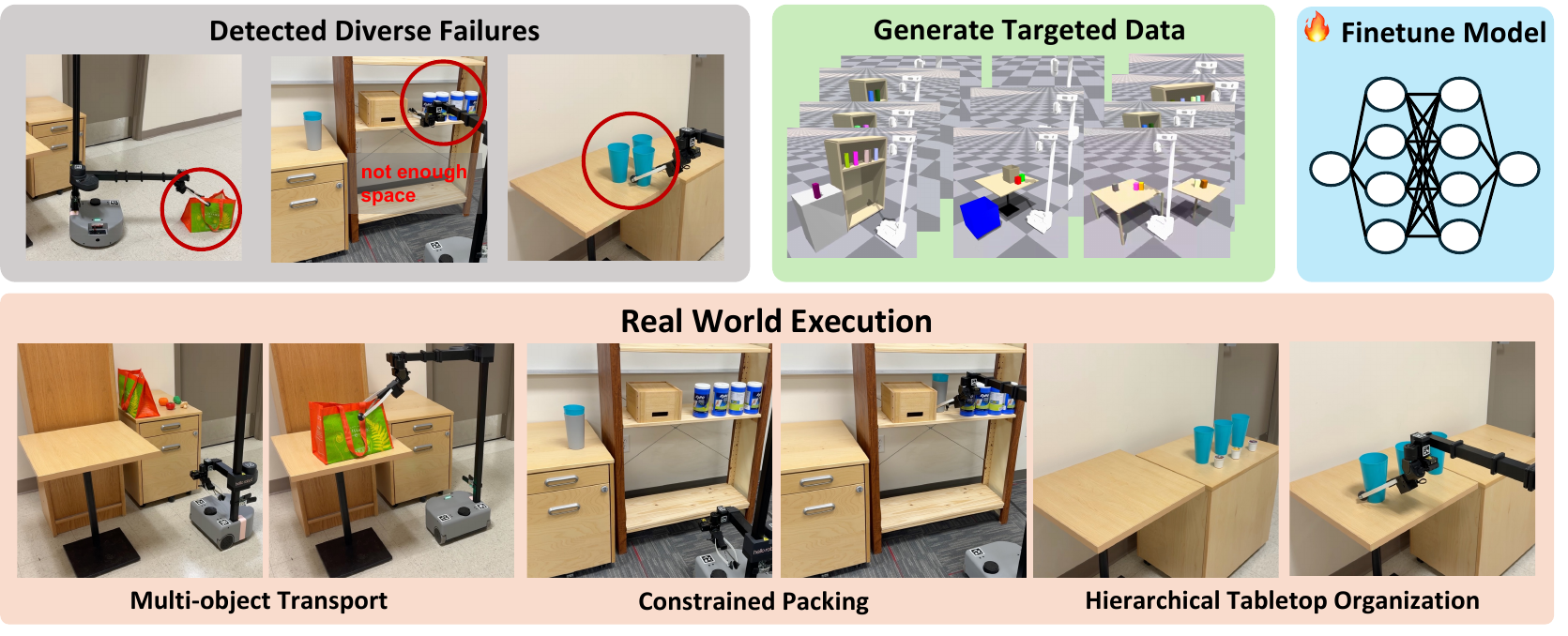} %
    \caption{\textbf{Overview of Failure Case Reasoning.} \textbf{Top}: Based on failure cases due to incorrect symbolic predictions, our approach generates targeted, diverse simulation data in parallel to fine-tune the robot's model. 
  \textbf{Bottom}: The fine-tuned model successfully performs diverse long-horizon manipulation tasks in challenging real-world scenarios. 
  \vspace{-10pt}
    }
    \label{fig:overview_figure}
    \vspace{-10pt}
\end{figure}
Skill-effects models predict the change in world state when running a skill given an initial observation and continuous parameters associated with the skill. These effects can be full metric states, such as the poses of objects, or symbolic states such as inter-object relations, logical states, or preconditions of other skills. 
Within this paradigm, we define a symbolic-level skill execution failure to occur when the world symbolic state after execution does not match the predicted (i.e.\ planned) symbolic effect state~(i.e.\ incorrect symbolic predictions). 
When operating with symbolic states, the robot does not need to perfectly match any predicted metric state as long as the high-level sub-goal is reached. 

Assuming the robot itself does not break and other agents do not disturb the environment, we can categorize failures into two types: (1) those arising from incorrect symbolic predictions~(e.g., Fig.~\ref{fig:overview_figure}) and (2) those resulting from a Sim-to-Real~(Sim2Real) gap during closed-loop skill execution.
Specifically, if the system achieves the desired symbolic outcome, it indicates there is no skill-symbolic Sim2Real gap, even if the robot acts somewhat differently in the real world than in the simulator. 
If a failure does occur, it arises either because the trained model would incorrectly predict effects in an equivalent simulation scenario or because the closed-loop execution in simulation deviates significantly from its real-world execution. 
The Sim2Real gap could be caused by real-world perception noise, controller mismatch, or inaccurate physical modeling in the simulation, among other causes~(details in Appx.~A.7). 
In this paper, we investigate detecting failures, classifying failure types, and learning from failures due to incorrect symbolic predictions. 
Note that our approach is complementary to other works~\cite{tremblay2018training, tobin2017domain, ramos2019bayessim} addressing the Sim2Real gap. 

While a robot could learn directly from real-world failure cases, a single failure instance is insufficient to effectively refine modern large parameter models~\cite{open_x_embodiment_rt_x_2023, khazatsky2024droid, li2014efficient}. 
The robot could instead try to explicitly generate more real-world failure scenarios~\cite{kumar2024practice, smith2024grow}, but making the robot explore in the open environment poses risks to the robot and the surrounding environment.  
To address this, Real-to-Sim~(Real2Sim) approaches~\cite{mandi2024real2code, chen2024urdformer, torne2024reconciling} have gained popularity in robot manipulation, as they enable safe and efficient creation of simulation environments. 
Nevertheless, current methods emphasize high-fidelity simulations, which can be computationally expensive and require extensive fine-tuning~\cite{mandi2024real2code, chen2024urdformer} or environment scanning~\cite{torne2024reconciling}. 
Therefore, generating diverse data conditioned on failure cases efficiently and safely to improve skill effect models is an important and open question to address. 

In this work, we advocate for generating low-fidelity simulation environments informed by real-world failures. 
We can generate such simulation datasets efficiently and safely to fine-tune skill effect models to minimize future failures in long-horizon tasks. 
Recent progress in physical simulation~\cite{isaacgym, mittal2023orbit} has shown success in accelerating simulation by running multiple environments in parallel on graphics processing units (GPUs). 
To leverage the power of parallel simulation, we propose to generate multiple simulation states in parallel. 
To this end, we formulate a variational inference problem to generate datasets targeted to observed failures for use in refining a skill effect model.
To efficiently generate samples in parallel, we propose using Stein variational inference~(\stein{})~\cite{pavlasek2023ready} as our variational solver. %

We introduce \svgd{}, which employs \stein{} to generate a simulation dataset informed by failure cases to enhance the skill effect model. When the robot detects a real-world failure occurrence, it records the relevant state information~(e.g., object relations), observation, and the executed robot skill associated with the failure. 
Given this, \svgd{} generates a simulation dataset that approximates the joint distribution of states that match the observed failure and actions that maximize the robot's information gain~\cite{smith2023prediction} of its current skill effect model. 
The robot then fine-tunes its skill effect model on this dataset to improve its future handling of scenarios similar to the observed failure, as shown in Fig.~\ref{fig:overview_figure}.

In summary, our contributions are:
(1): the formulation of a two-category failure classification problem between incorrect symbolic predictions and a Sim2Real gap. 
(2): We are the first to formulate failure case reasoning for long-horizon manipulation tasks as a variational inference problem, enabling the generation of diverse simulation data to improve the skill effect model. 
(3): We propose using Stein variational inference to approximate multi-modal posterior distributions over simulation states and robot skill parameters, thereby facilitating effective and efficient dataset generation. 
(4):~We implement our proposed approach with three distinct skill effect model formulations~\cite{huang2024points2plans, chen2023predicting, paxton-corl2021-semantic-placement}, efficiently generating low-fidelity simulation environments. 
Through large-scale mobile manipulation experiments, we demonstrate that \svgd{} outperforms several baselines.

\section{Related Work}
Failure case reasoning has gathered significant attention in the robotics community~\cite{farid2022failure, inceoglu2021fino, agia2024unpacking, sharma2021sketching, antonante2021monitoring, vats2023efficient, namasivayam2024learning, vats2024recoverychaining}. 
However, these works focus on detecting failures and recovering from failures, but do not explicitly address the problem of learning to minimize future failures.
Thomason and Kress-Gazit~\cite{thomason2023counterexample} propose to automatically improve a symbolic abstraction of a robot skill from observed failures. In contrast, we focus on improving pretrained skill effect models. 
Kumar et al.~\cite{kumar2024practice} enable the robot to explore real-world environments to improve its model. 
While effective, their approach is limited to closed environments and faces safety and efficiency challenges when generalizing to open, unstructured settings.

Real-to-Sim approaches show promise in efficiently generating large-scale simulation datasets for policy learning~\cite{torne2024reconciling, torne2024robot, wang2023real2sim2real, chen2024urdformer, lim2021planar}, and improving simulation physical parameters using real-world data~\cite{memmel2024asid, chebotar2019closing, ma2023sim2real, ramos2019bayessim, antonova2022bayesian}. 
Additionally, there is extensive literature on generating articulated objects from real-world 2D or 3D data~\cite{qian2022understanding, chen2024urdformer, mandi2024real2code, jiang2022ditto, heiden2022inferring, mao2022multiscan}.
However, these approaches primarily focus on creating high-fidelity simulations, which can be computationally inefficient. 

Stein variational inference~(\stein{}) has found application in robotics due to its ability to approximate high-dimensional, multimodal posterior distributions~\cite{pavlasek2023ready, Barcelos-RSS-21, lee2023stamp, power2024constrained, lambert2021entropy, lambert2022stein, honda2024stein}. Existing use cases focus on generating diverse and robust plans under uncertainty in various contexts~\cite{Barcelos-RSS-21,lambert2022stein, pavlasek2023ready,lee2023stamp}.
We instead use \stein{} to generate diverse simulated datasets to improve learned models. 

\section{Skill Effect Models}
We build our failure reasoning approach on top of skill effect models~\cite{huang2024points2plans, chen2023predicting, paxton-corl2021-semantic-placement} that can solve long-horizon, geometrically complex tasks directly from high-dimensional, partial-view point clouds. 
Our skill effect models require segmented point clouds and thus assume a perception pipeline with (1): the semantics of objects and (2): the segmentation of each object. 
In this paper, we use open-source models for segmentation~\cite{kirillov2023segment} and detection~\cite{liu2023grounding}. 
We assume a given set of manipulation skill primitives such as push, pick and place, etc. $\mathcal{L} = \{\phi^1, ..., \phi^K\}$. 
Each primitive $\phi^k$ is parameterized with a continuous parameter $a^k \in \mathbb{R}^m$.
A robot skill $\phi^k(a^k)$ is a skill primitive parameterized by $a^k$ that can be executed on the robot. 
A set of relations $\mathcal{R} = \{\texttt{on}, \texttt{left}, \texttt{right}, ...\}$ is also given, the relations include inter-object relations or single-object relations. 
Inter-object relations include spatial relations like \texttt{left} and physical relations like \texttt{in-contact}. 
Single-object relations include whether an object is \texttt{manipulable} ~(e.g., a shelf is not manipulable) and whether a drawer is \texttt{open}. 
The plan skeleton $\phi_{1:H}$ is defined as a sequence of feasible primitives.  
When the plan skeleton is paired with valid continuous parameters, it enables the robot to achieve the goal relations from the initial state.  

A skill effect model can be trained with a large-scale simulation dataset $\dataset = \{(s_t, O_t, \phi_t, a_t, s_{t+1}, O_{t+1})\}$, where $O_t$ represents the observation at time $t$ as segmented point clouds and $s_t$ represents the simulation state information including geometric information like object pose and physical information like the object friction parameters and $\phi_t$ represents the skill to execute on the simulated robot with corresponding continuous parameters $a_t$. 
Given the initial simulation state $s_t$, we can get the effects of the skill $\phi_t(a_t)$. 
The ground-truth relations $r_t$ are a function of the simulation state $s_t$. 
We train the model, $\Gamma$, with the dataset, $\dataset$. 
The trained model can predict the probability of achieving specific relations based on an initial point clouds observation, $O_0$, a robot skill sequence, and a training dataset as $\Gamma(r | O_0, \phi_{1:H}, a_{1:H}, \dataset)$, where $r \in \mathcal{R}$. 
Given a goal, $\mathcal{G}$, defined as a conjunction of desired relations $g_1 \land ... \land g_M$, $g_i \in \mathcal{R}$.  The planning objective is to find a skill sequence $\phi_{1:H}(a_{1:H})$ that maximizes the probability that the goal relations are achieved $p(\mathcal{G} | O_0, \phi_{1:H}, a_{1:H}, \dataset)$, where the plan skeletons $\phi_{1:H}$ could be generated using any number of techniques such as: foundation models~\cite{huang2024points2plans, lin2023text2motion}, graph search~\cite{huang-erd-tro2024}, or other classical planners~\cite{garrett2021integrated}. 
Given the skeleton, one can maximize the planning objective using standard numerical optimization techniques such as a shooting method~\cite{huang2024points2plans, stap, lin2023text2motion} or a cross-entropy method~\cite{huang-erd-tro2024, huang-icra2023-graph-relations, huang-icra204-memory}) to generate continuous parameters $a_{1:H}$.
For more details~(e.g., implementation details) about the skill effect model, please refer to Appx.~A.11.%

\section{Detecting and Classifying Failures Autonomously}\label{sec:failure-detection}
Consider a robot operating using a skill effect model $\rd$ trained on some dataset, \(\dataset\).
A user tasks the robot to achieve a goal, $\mathcal{G}$.
Given an initial observation, $O_0$, its skill effect model, and the goal, the robot plans a skill sequence, $\SkillSeq$. In addition to the skill sequence itself the skill effect model predicts a sequence of expected relations, \(\mathcal{R}_{k}^{\prime}\) (i.e. symbolic states) that the robot will observe after executing each skill in its sequence for \(k=1,\ldots,H\).
The robot now begins executing its skills following the plan.
After each skill execution in the sequence, the robot can observe the current scene, \(O_k\), and detect the current symbolic state as \(\hat{\mathcal{R}}_k\). When the observed relations, \(\hat{\mathcal{R}}_k\), don't match the predicted relations, \(\mathcal{R}_k^{\prime}\), the robot detects that it has failed to achieve the current subtask. 

In the case of a detected failure, the robot stores the associated failure event in order to learn from it.
We define a failure event to include the relevant observations, relations (i.e. symbolic states), and skill associated with the failure: \(\fail = (O^{\fail}=O_{k-1}, \mathcal{R}^{\fail}=\hat{\mathcal{R}}_{k-1}, \phi^{\fail}=\phi_k, \mathcal{R}^{\fail\prime}=\hat{\mathcal{R}}_{k})\).
Once the robot has detected a failure, it will classify the failure category. It first reconstructs the same simulation based on observation $O^{\fail}$ and predicts the relational effects of the action $\phi_k(a_k)$ as $\mathcal{R}_k^{\prime\prime}$. If the simulation relational effects $\mathcal{R}_k^{\prime\prime}$ match the real-world relational effects $\mathcal{R}_k^{\prime}$, then the robot will classify this failure as stemming from incorrect symbolic predictions. Otherwise, this failure is classified as a Sim2Real gap. We now consider the problem of learning from the failure instance to improve the skill effect model $\rd$. 

\section{Generating Targeted Datasets to Learn from Failure} \label{sec:approach}
Since modern, large neural networks typically cannot learn from a single failure instance~\cite{open_x_embodiment_rt_x_2023, khazatsky2024droid, li2014efficient}, we pose the problem of learning from failures as a problem of efficiently collecting a new dataset $\dataset^+$. Determining which data points from an infinite possible set to generate and label can naturally be defined as an active learning problem~\cite{settles2012active,conkey-humanoids2019-active-promp,lu-iros2020-active-grasp, smith2023prediction}. 
From this perspective, we can quantify the effectiveness of the generated additional training dataset $\dataset^{+}$ using the expected information gain criteria~\cite{smith2023prediction}. 
However, we want to target our new dataset to be similar to the scenario in which the robot failed, something which information gain alone does not address. 

We thus define our problem as finding a dataset $\dataset^+$ that yields high expected information gain in terms of improvement in the predictions from $\rd$ associated with the detected failure.
At the same time, we ensure that the samples in \(\dataset^{+}\) have a high probability of the same relations observed by the robot prior to executing the failed skill.
Let us first define the form of a single sample \(d_i^{+} \in \dataset^{+}\), as \(d_i^+ = (s_i^+, O_i^+, \phi^{\fail}, a_i^{+}, s_{i}^{++}, O_i^{++})\). Note that the skill, \(\phi^{\fail}\), is fixed for all samples to be the skill that failed to achieve the subtask. Next let us define the results of evaluating the sample action \(a_i^{+}\) in the simulator $f$ as \(s^{++}_i = f(s_i^{+}, \phi^{\fail}, a_i^{+})\), which when rendered defines the post skill observation \(O_{i}^{++} = \render(s^{++}_i)\). We also render pre-action states to observations as \(O_i^+ = \render(s_i^+)\). Thus, samples in \(\dataset^{+}\) have only two free variables for us to search over: the initial simulator state, \(s^+_i\); and the action to execute \(a_i^{+}\). We will use $S^{+}$ to denote the set of state samples, \(\{s^+\}\) in \(\dataset^+\), and \(S\) to denote the random variable associated with the state. We will use a similar notation for actions and observations.

We can formalize our dataset generation problem as the following constrained optimization problem, noting that maximizing the expected information gain is equivalent to maximizing the KL-divergence between the predictive distributions of the updated model $\Gamma^{+}$ fine-tuned on \(\dataset^{+}\), and the original model $\Gamma$
\begin{subequations}\label{eq:KL-problem}
\begin{align}
  \argmax_{\dataset^{+}} \quad &D_{KL}\left(\prod_{r \in \mathcal{R}^{\fail\prime}} \rd^{+}(r | O, \phi^{\fail}, A, \dataset \cup \dataset^{+}) \; \bigg\| \;  \prod_{r \in \mathcal{R}^{\fail\prime}}\rd(r | O, \phi^{\fail}, A, \dataset)\right) \label{eq:KL-Obj} \\
    \text{subject to} \quad & S^+ \sim P(\mathcal{R}^{\fail}, O^{\fail}|S) \label{eq:KL-constraint}
\end{align}
\end{subequations}
Thus, we must find a set of simulator states, \(S^+\), and actions \(A^+\) which maximize the active learning objective, while also ensuring the sample states would generate the same relations and point cloud observations observed by the robot before the failure. This distribution in Eq.~\ref{eq:KL-constraint} factorizes as 
\begin{equation}
P(\mathcal{R}^{\fail}, O^{\fail}|S) = \prod_{r \in \mathcal{R}^{\fail}} \rd(r|O = \render(S)) P(O^{\fail}|S) P(S) \label{eq:obs-dist}
\end{equation}
where the first term, $\rd(r|O = \render(S))$, encodes the objective of finding states in the simulator that achieve the same relations when rendered and evaluated by the skill effect model. 
The second term, \(P(O^{\fail}|S)\), ensures that we generate point clouds that match the failure observation. The final term, \(P(S)\), encodes a prior over valid states in the simulator.

This formulation presents several computational challenges. (1) The objective in Eq.~\ref{eq:KL-Obj} is intractable because there exists an infinite number of possible datasets, $\dataset^{+}$.  %
(2) Evaluating Eq.~\ref{eq:KL-problem} requires running the simulator to generate all samples in the putative \(\dataset^{+}\) and retraining the skill effect model $\Gamma$ for each possible dataset. %
(3) Finding simulator states \(s_i^{+}\) that render to point clouds matching the failure observation, amounts to an inverse problem over object geometries and poses. 
(4) Ensuring that the states obey the constraint while maximizing the objective defines a high-dimensional, non-convex problem.

\subsection{Approximate Constrained Expected Information Gain}
We propose two specific approximations to the problem defined in Eq.~\ref{eq:KL-problem} in order to make the problem tractable.
We can summarize these approximations in the following problem:
\begin{subequations}\label{eq:approx-problem}
\begin{align}
    \argmax_{{S}^{+}, {A}^+} \, &\prod_{r \in \mathcal{R}^{\fail\prime}}H\left(\rd(r\mid\xi(S)O^{\fail}, \phi^{\fail}, A,\dataset)\right)\label{eq:approx-objective}\\
    \text{subject to} \quad & S^+ \sim \rd(r^{\fail}\mid O = \xi(S)O^{\fail})P(S) \label{eq:approx-constraints}
\end{align}
\end{subequations}
Here we have replaced the expected information gain objective in Eq.~\ref{eq:KL-Obj} with the entropy defined over the epistemic uncertainty~\cite{hullermeier2021aleatoric, kendall2017uncertainties} of the currently trained model, \(\rd(\cdot, D)\), where \(H(P(Y\mid X)) = -\sum_{y\in Y} P(Y=y\mid X) \ln{P(Y=y \mid X)}\).
As a common approximation widely used in active learning~\cite{settles2012active,conkey-humanoids2019-active-promp}, this allows us to avoid running the simulator and fine-tuning at each iteration of dataset optimization, thereby simplifying the objective. 
The distribution defined in Eq.~\ref{eq:obs-dist} implies that one needs to search over object poses and geometries that match the appearance of the partial-view point clouds observed during the failure event. 
This defines an infinitely large space of possible object shapes, which we wish to avoid searching over. 
Instead, we simplify the constraint to transpose the poses of the individual object point clouds in \(O^F\), while ensuring that the point clouds still achieve the same relations when evaluated by the detector, i.e. \(\rd\) evaluated without any actions. 
We denote by \(\xi(s)O^{\fail}\) the segment-wise transformation of the point cloud to the poses defined by the state vector \(s\). Note, this allows us to search over object poses without using the full physics simulator or renderer. We use the simulator to generate \(\dataset^+\) after finding \(S^{+}\) and \(A^{+}\). We describe how we instantiate this transformed point cloud to a full object for the simulator in Sec~\ref{sec:real2sim}.

\begin{figure}[th]
    \centering
    \includegraphics[width=0.99\columnwidth]{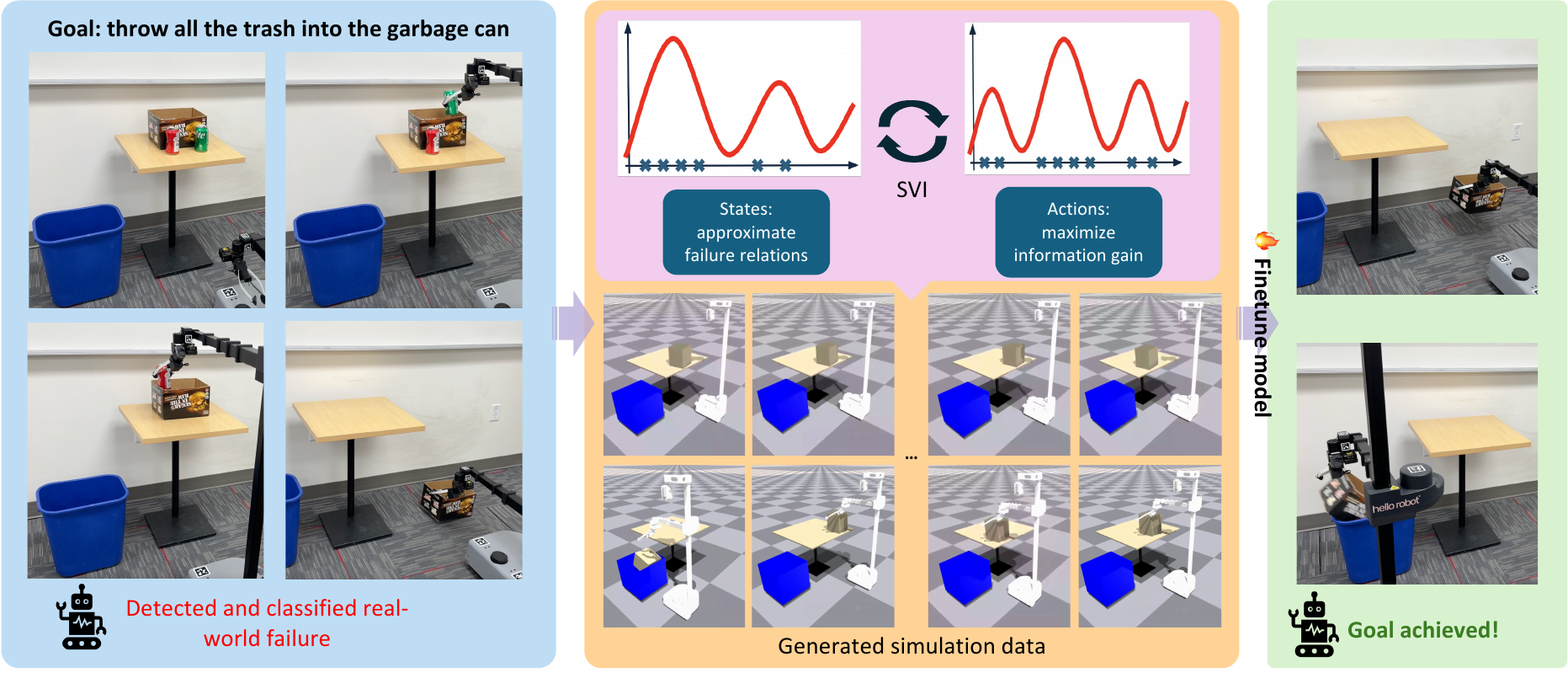}
    \caption{Overview of \svgd{}. Our approach first detects the real-world failure and classifies it as incorrect symbolic predictions. 
    Based on the real-world failure, \svgd{} generates particles representing simulation states and actions to approximate posterior distributions. 
    The state posterior distribution captures failure relations~(e.g., objects inside the box), while the action posterior distribution maximizes the information gain of the skill effect model. 
    Using these particles, a diverse simulation dataset is created to fine-tune the skill effect model. 
    After fine-tuning, the model successfully recovers from the failure and completes the real-world task of cleaning.\vspace{-10pt}}
    \label{fig:approach}
\end{figure}

\subsection{Generating Datasets via Stein Variational Inference}
We approximately solve the constrained optimization in Eq.~\ref{eq:approx-problem} in two stages.
First, we find a set of state samples $S^{+}$ that approximates the posterior distribution defined by Eq.~\ref{eq:approx-constraints}. We formulate finding this set as a variational inference problem.
Then keeping $S^+$ fixed, we solve for the continuous action parameters $A^+$ that maximize Eq.~\ref{eq:approx-objective} using generalized Bayesian inference~\cite{matsubara2022robustGBI,pavlasek2023ready}. 
To solve both inference problems, we leverage Stein variational gradient descent which we now review.

\textbf{Stein Variational Gradient Descent: }
In variational inference one defines a tractable distribution \(q(X)\) to approximate the target distribution \(P(X)\)~\cite{Blei_2017}. One then optimizes over the parameters defining the variational distribution \(q\) in order to minimize the KL-divergence between the variational distribution and the target distribution \(P(X)\) as $\underset{q(X)}{\arg\min} \dkl{q(X)}{P(X)}$. 
Stein variational inference represents the posterior as a set of particles \(q=\{x_i\}_{i=1}^M\). 
Stein variational gradient descent~\cite{liu2016stein}~(SVGD) leverages gradient-based optimization to guide the particles in a direction that minimizes the KL divergence. SVGD performs efficient approximate inference through parallel gradient-based optimization and can contend with high-dimensional and multi-modal posterior distributions. An SVGD particle $x_i$ is updated at iteration $k$ as $x^{+(k)}_i \leftarrow x^{+(k-1)}_i + \eta \Phi({x^{+(k-1)}_i})$,  
where $\Phi$ is the Stein variational gradient computed using the Stein operator and a kernel $k(x^{+}_{j}, x_i^{+})$. In this paper, we use a radial basis function (i.e. squared-exponential) kernel and set its kernel bandwidth using the median heuristic~\cite{pavlasek2023ready, garreau2017large}.

In the case of generalized Bayesian inference (GBI), we modify the variational inference objective to account for defining a loss function over our variables instead of a traditional likelihood~\cite{matsubara2022robustGBI,pavlasek2023ready}. Given a loss function \(\mathcal{L}(Y,X)\) for an arbitrary random variable \(X\) and some observations \(Y\), GBI defines the following approximate posterior $P_{\mathcal{L}}(X\mid Y) \propto P(x)\exp \left( -\beta \mathcal{L}(Y, X) \right)$. 
We can then use this approximate posterior within a Stein variational inference framework by solving the following problem $\underset{q \in \mathcal{Q}}{\arg\min} \beta \; \meandist{\mathcal{L}(X, Y)}{X\sim q} + \dkl{q(X)}{P(X)}$.

\textbf{Generating State Samples: }
We want to find samples $q(S) = \{s_i^{+}\}_{i=1}^M$ that approximate the posterior distribution $P(r^{\fail}\mid O^{+} = \xi(S)O^{\fail})P(S)$, where $P(S)$ is a uniform prior over all feasible states. 
The posterior distribution ensures that the transformed point clouds match the relations in the failure case $r^{\fail}$. 
This defines the following variational inference problem: $\underset{q(S)}{\arg\min} \dkl{q(S)}{\rd(r^{\fail}\mid O^{+} = \xi(S)O^{\fail})P(S)}$ and it's solved by using \stein{}. 

\textbf{Generating Action Samples: } 
Given our state samples generated using Stein variational inference to approximate the distribution in Eq.~\ref{eq:approx-constraints}, we can now turn our attention to solving for the action set \(A^{+}\). To formulate this problem we make use of the generalized Bayesian inference framework outlined above. Here we define the loss function, \(\mathcal{L}\), to be the entropy loss defined in Eq.~\ref{eq:approx-objective} and let \(\beta=1\). The variational distribution takes the form \(q(A) = \{(s^{+}_i, a^{+}_i)\}_{i=1}^M\), where we keep the values of \(s_i^{+}\) fixed and search only over actions. This defines the following variational inference problem: $\underset{q(A)}{\arg\min}\, \meandist{\prod_{r \in \mathcal{R}^{\fail}} -H(\rd(r\mid \xi(s^{+})O^{\fail}, \phi^{\fail}, a^{+}, D))}{s^+, a^+ \in q(A)} + \dkl{A^{+}}{P(A)}$, 
where $P(A)$ is a uniform prior over actions. 
We solve the variational inference problem for generating state and action samples using \stein{}. We provide the details in Appx.~A.13. %

\subsection{Real-to-Sim Object Generation}
\label{sec:real2sim}
After running the two Stein inference procedures, we obtain optimized particles denoted as $\{s^{+}_i, a_{i}^{+}\}_{i=1}^M$. We now must generate the full dataset in simulation. %
First, we generate a simulation scene based on $s^{+}_i$ and then execute the corresponding robot skill, $a_{i}^{+}$. 
Since a failure observation, $O^{\fail}$, contains semantic segments for each object, we fit the corresponding object shapes~(e.g., cuboids, open boxes, and drawers) based on these semantics. 
The bounding box of each segment determines the size of each object and combined with each object's pose, we construct the simulation scene. 

We use pre-defined physical parameters~(e.g., friction and center of mass) for each object class. 
After creating the simulation scene using $s^{+}_i$ and executing robot skill, \((\phi^{\fail}, a_{i}^{+})\), we obtain our fine-tuning dataset, $\dataset^{+}$, to refine the skill effect model $\rd$. 
We note that our bounding-box-based real-to-sim scene generation is chosen primarily for its efficiency,  allowing us to compare different dataset generation approaches. 
However, our primary contribution, \svgd{}, is complementary to other real-to-sim methods~\cite{mandi2024real2code, torne2024reconciling, chen2024urdformer}.
If simulated objects generated by a Real2Sim pipeline do not capture important features of real objects, they may cause (1): incorrect classifications of failure types~(incorrect symbolic predictions vs Sim2Real gap), and (2): poor fine-tuning data generation to improve the model. We examine the quality of our Real2Sim approach in the context of \svgd{} in Sec.~\ref {sec:exp}.

\section{Experiments \& Results}\label{sec:exp}
In this section, we present both simulation and real-world experiments to address several key questions:
(Q1): To what extent does learning from failures improve long-horizon manipulation tasks? 
(Q2): Is adding more data sufficient to resolve all failure cases? 
(Q3): Can replanning alone effectively handle all failure scenarios? 
(Q4): How does \svgd{} perform under noisy input? 
(Q5): Does \stein{} improve the performance of \svgd{}? 
(Q6): Can \svgd{} generalize to novel scenarios absent from both the pre-training and fine-tuning datasets? 
(Q7): Is Real2Sim accurate enough for our tasks? 

We first use IsaacGym~\cite{isaacgym} to generate a pre-training dataset containing 40,000 skill executions, which is larger than the datasets used in~\cite{huang2024points2plans, chen2023predicting, paxton-corl2021-semantic-placement}. 
After pre-training a skill effect model with this dataset, it is evaluated across diverse scenarios until a failure is detected. 
Based on the observed failure, \svgd{} utilizes \stein{} to generate particles representing simulation states and corresponding robot actions. 
Once the simulation scenes are created and the robot skills are executed, \svgd{} generates a targeted simulation dataset to fine-tune the skill effect model. 
Specifically, this targeted dataset consists of 20 diverse environments and their corresponding robot actions. 
We chose 20 data points based on an ablation study presented in Appx.~A.5. %
Figure~\ref{fig:approach} visualizes an example simulation dataset generated by our approach.

\textbf{Baseline Comparisons: }
In this paper, we compare our proposed approach, \svgd{}, against six baselines on three state-of-the-art skill effect models~\cite{huang2024points2plans, chen2023predicting, paxton-corl2021-semantic-placement}:  
\textbf{\orig{}~\cite{huang2024points2plans, chen2023predicting, paxton-corl2021-semantic-placement}:} These are the state-of-the-art skill effect models without fine-tuning. It serves as a baseline to demonstrate the performance of pure skill effect models without learning from out-of-distribution failures. 
\textbf{\sma{}:} This baseline was trained using the pre-training dataset with a small set~(40000+20 data points). 
\textbf{\lar{}:} This baseline used the pre-training dataset alongside a much larger dataset~(40000+2000 data points). 
\textbf{\gradient{}~\cite{lu-iros2020-active-grasp}:} This approach uses stochastic gradient descent to update individual particles, employing the same objective as \svgd{}, while 
generating each sample sequentially and independently. 
This baseline serves to demonstrate the effectiveness of \svgd{}. 
\textbf{\sampling{}~\cite{conkey-humanoids2019-active-promp}:} This sampling-based method iteratively generates each simulation state with the corresponding robot action until they satisfy the objective~\cite{gilks1992adaptive}. 
We use this baseline to highlight the challenges faced by sampling-based approaches in high-dimensional spaces. 
\textbf{\replan{}~\cite{liu2023reflect}:} This approach relies exclusively on replanning to recover from failure cases. 
We include it to assess whether a pure replanning strategy is sufficient for failure recovery. 
We evaluate our approach against baselines in the tasks shown in Fig.~\ref{fig:overview_figure}. We provide detailed explanations of these tasks in Appx.~A.2.%

\begin{wraptable}{rt}{0.6\textwidth}
    \vspace{-10pt}
    \centering
    \caption{\small
        Simulation experiments for the \textbf{Hierarchical Tabletop Organization}. 
   \svgd{} outperforms all baselines by a large margin. 
    Further details are reported in Appx.~A.6.%
    \vspace{-12pt}
    }
    \label{table:simulation}
    \adjustbox{max width=\textwidth}{
        \begin{tabular}{lccccccc}
            \toprule
            & \orig{} & \sma{}  & \lar{}  & \replan{}  & \sampling{} & \gradient{} & \svgd{}\ \\
            \midrule
            \point{}~\cite{huang2024points2plans} & 11\% & 13\% & 16\% & 24\% & 53\% & 45\% & \textbf{86\%} \\
            Stow-GNN~\cite{chen2023predicting} & 10\% & 11\% & 15\% & 21\% & 51\% & 44\% & \textbf{80\%} \\
            Binary-Pred~\cite{paxton-corl2021-semantic-placement} & 8\% & 9\% & 12\% & 19\% & 41\% & 38\% & \textbf{72\%} \\ \bottomrule
        \end{tabular}
    }
    \vspace{-10pt}
\end{wraptable}
\textbf{Simulation Success Rate Evaluation:} 
We first compare \svgd{} against all baselines on the \textbf{Hierarchical Tabletop Organization} task.  
We evaluate execution success rates across varying numbers of objects, as shown in Table~\ref{table:simulation}. 
Each approach is tested with 300 trials. 
The comparison between \svgd{} and \orig{} demonstrates that \svgd{} significantly outperforms \orig{}, highlighting the importance of learning from out-of-distribution failures~(Q1). 
The comparison shows that \svgd{} outperforms \sma{} and \lar{} by a large margin, demonstrating that merely adding more data is insufficient to resolve all failure cases, highlighting the importance of our approach in selecting quality, targeted data~(Q2). 
While \replan{} performs slightly better than \orig{}, it falls far short of \svgd{}, demonstrating that replanning alone is insufficient for recovering from failures, particularly in scenarios with large prediction errors in the dynamics model~(Q3). 
Furthermore, the success rate comparisons among \svgd{}, \gradient{}, and \sampling{} reveal that \svgd{} consistently outperforms both baselines~(Q5). 
This superior performance is attributed to \stein{}'s capability to effectively approximate high-dimensional, multi-modal posterior distributions. 
Furthermore, the average relation detection F1 score of our approach is 0.92, ensuring reliable failure detection. 

\begin{wraptable}{rt}{0.5\textwidth}
    \vspace{-10pt}
    \centering
    \caption{\small
        We show the generalization capability of \svgd{} in simulation across varying numbers of objects and different viewpoints. The \btext{5objs, 7objs, view1, and view2} scenarios are unseen during training.
        \vspace{-10pt}
    }
    \label{table:generalization_main}
    \adjustbox{max width=\textwidth}{
        \begin{tabular}{lccccc}
            \toprule
            Generalization Scenarios & 3objs $\uparrow$ & \bcl{5objs} $\uparrow$ & \bcl{7objs} $\uparrow$ & \bcl{view1} $\uparrow$ & \bcl{view2} $\uparrow$\\ \\
            \midrule
            \svgd{}  &  \textbf{87\%} & \textbf{81\%} & \textbf{71\%} & \textbf{83\%} & \textbf{85\%}\\
             \gradient{}  &  51\% & 40\% & 18\% & 42\% & 44\%\\
             \sampling{}  &  62\% & 45\% & 23\% & 51\% & 47\%\\
            \bottomrule
        \end{tabular}
    }
    \vspace{-10pt}
\end{wraptable}
\textbf{Generalization Evaluation: }
We assess the generalization capability of \svgd{} compared to \gradient{} and \sampling{} in the \textbf{Multi-object Transport} task with the \point{} architecture. 
We evaluate generalization to an unseen number of objects and unseen viewpoints shown in Table~\ref{table:generalization_main}. 
While all approaches experience some performance degradation, \svgd{} maintains strong performance~(Q6).

\begin{wrapfigure}{rt}{0.6\textwidth}
    \vspace{-15pt}
    \centering
   \includegraphics[width=1.0\textwidth,clip,trim=0mm 0mm 0mm 0mm]{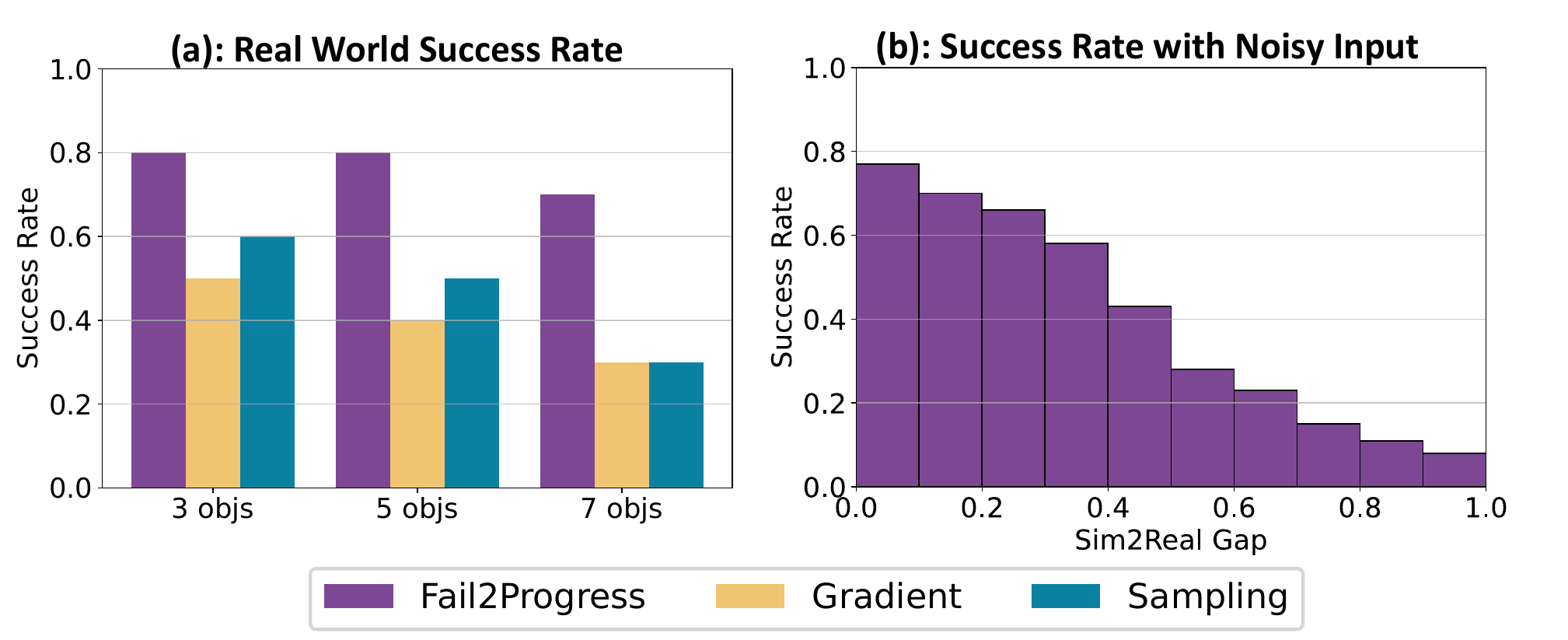} %
    \caption{\small
    \svgd{} consistently achieves higher success rates than all baselines on the \textbf{Hierarchical Tabletop Organization} task~(a).   
    Without artificially added noise in the point clouds, \svgd{} performs well. However, its performance degrades under a large Sim2Real gap caused by noisy point clouds~(b).  
    \vspace{-10pt}
    }
    \vspace{-10pt}
    \label{fig:real}
\end{wrapfigure}
\textbf{Real-world Quantitative Evaluation: }~\label{sec:real}
We compare our approach, \svgd{}, with the \gradient{} and \sampling{} baselines in real-world scenarios with the \point{} architecture.
As shown in Fig.~\ref{fig:real}a, \svgd{} consistently outperforms both baselines~(Q5). 
For the real-world experiments, we performed 10 trials per approach for each object count. 
We also examine how the Sim2Real gap affects the performance of \svgd{}, as shown in Fig.~\ref{fig:real}b.  
Without artificially added noise in the real-world point clouds, \svgd{} successfully detects failures, classifies the failure reason as incorrect symbolic predictions, and achieves a high success rate shown in the left bar of Fig.~\ref{fig:real}b, which demonstrates that Sim2Real gap is small and Real2Sim is accurate enough~(Q7). 
However, the Sim2Real gap becomes a significant issue when the real-world point clouds are noisy. 
To analyze this effect, we add random Gaussian noise to the point clouds. 
We quantify the Sim2Real gap as the difference in predicted relations between the real-world and reconstructed simulation scenarios, using the same robot skill. 
With noisier point clouds, the Sim2Real gap increases, leading to a degradation in the performance of \svgd{}~(Q4). 
We provide efficiency experiments~(Appx.~A.3), further qualitative analysis~(Appx.~A.1) and a summary of key findings~(Appx.~A.4) in the appendix.

\section{Conclusion}
This work addresses the problem of learning from failures in long-horizon manipulation tasks using learned skill effect models. 
We propose generating additional, targeted simulation datasets based on observed failures to fine-tune the pre-trained skill effect model. 
We formalize the task as a probabilistic inference problem that maximizes the information gain of the datasets while ensuring the datasets remain close to the observed failure. 
To solve it, we introduce \svgd{}, an approach that leverages \stein{} to approximate multi-modal posterior distributions. 
Through experiments, we demonstrate that \svgd{} can generate failure-driven simulation datasets to improve the skill effect model more effectively and efficiently compared to six baselines. 
Furthermore, we deploy \svgd{} on a mobile manipulator, showcasing its ability to perform diverse real-world tasks, such as packing groceries, packing a constrained shelf, and organizing a table. 

\section{Limitations}
Our approach has several limitations. 
First, although \svgd{} significantly improves performance, it still falls short of perfect reliability, achieving around an 80\% success rate in the real world shown in Fig.~\ref{fig:real}a.
This is because, even after fine-tuning, some scenarios remain out-of-distribution, leading to incorrect symbolic predictions. Indeed, one can think of the results presented in this paper as "1-shot" \svgd{} and that further refinement on the observed failures would lead to higher future success rates.
To continuously improve the performance as a lifelong learning system, the framework needs to be deployed in a real environment over several days, where we allow \svgd{} to update as needed when failures are detected and classified as being caused by incorrect symbol predictions.
Safely deploying \svgd{} in such open environments remains an open research question. 
Furthermore, our framework needs to be evaluated under more diverse conditions, including more complex and dexterous manipulation tasks involving varied objects, such as deformable objects and liquids. 

Second, we do not investigate correcting for failures caused by the Sim2Real gap in this work. 
The Sim2Real gap could potentially be mitigated by methods that explicitly address this challenge~\cite{tremblay2018training, tobin2017domain, ramos2019bayessim}. Showing how to integrate Sim2Real improvements alongside symbolic prediction failures is an important next step.

Third, we rely on Real2Sim to classify failures and generate high-quality fine-tuning datasets. 
Though our experiments show that our Real2Sim solution is effective in classifying failures and improving model performance, our Real2Sim itself is not perfect, especially when modeling complex object geometries and deformable objects. 

Fourth, our failure classification scheme, which includes two categories, does not explicitly reason about the environmental disturbances caused by other agents~(human users or other robots). It additionally does not account for hardware breaking or changing over time (e.g., cable or belt stretch in a robot arm drivetrain), which might occur over long deployment times.
Hypothesizing these scenarios as failure causes is also an interesting future direction. 

Fifth, we consider only object poses as the simulation state. Incorporating additional simulation states, such as object friction and center of mass~\cite{memmel2024asid}, into our framework would be a possible next step. 

Sixth, we assume a fixed set of relations. While our large-scale experiments show that these relations are sufficient, there are always relations outside the predefined set. 
Discovering new relations~\cite{shah2024reals, ahmetoglu2024discovering} during robot exploration could enhance the open-world planning capability of our framework. 

Finally, although we demonstrate mobile manipulation in diverse environments, extending the system to building-wide open spaces~\cite{shah2024bumble} remains an open research question.
To achieve this, our method could integrate with scene graph construction and online updating~\cite{liu2024dynamem, tang2025openin, rana2023sayplan, agia2022taskography}.

\section*{Acknowledgments} 
We would like to thank Jeannette Bohg and Christopher Agia for fruitful discussions. 
This work was partially supported by NSF Awards \#2149585, \#1846341, and \#2321852, by DARPA under grant N66001-19-2-4035, and by a Sloan Research Fellowship.

\bibliography{references}

\newpage
\pagenumbering{arabic}%
\renewcommand*{\thepage}{A\arabic{page}} %
\appendix
\section{Appendix}
\textbf{Overview}

The appendix provides additional details, experiments, and results. 
Please refer to the supplemental video for real-world robot executions available at \href{https://sites.google.com/view/fail2progress}{sites.google.com/view/fail2progress}. 
\startcontents[sections]
\printcontents[sections]{l}{1}{\setcounter{tocdepth}{2}}
\newpage
\subsection{Qualitative Analysis}\label{sec:qualitative}
\begin{figure}[th]
    \centering
    \includegraphics[width=0.99\columnwidth]{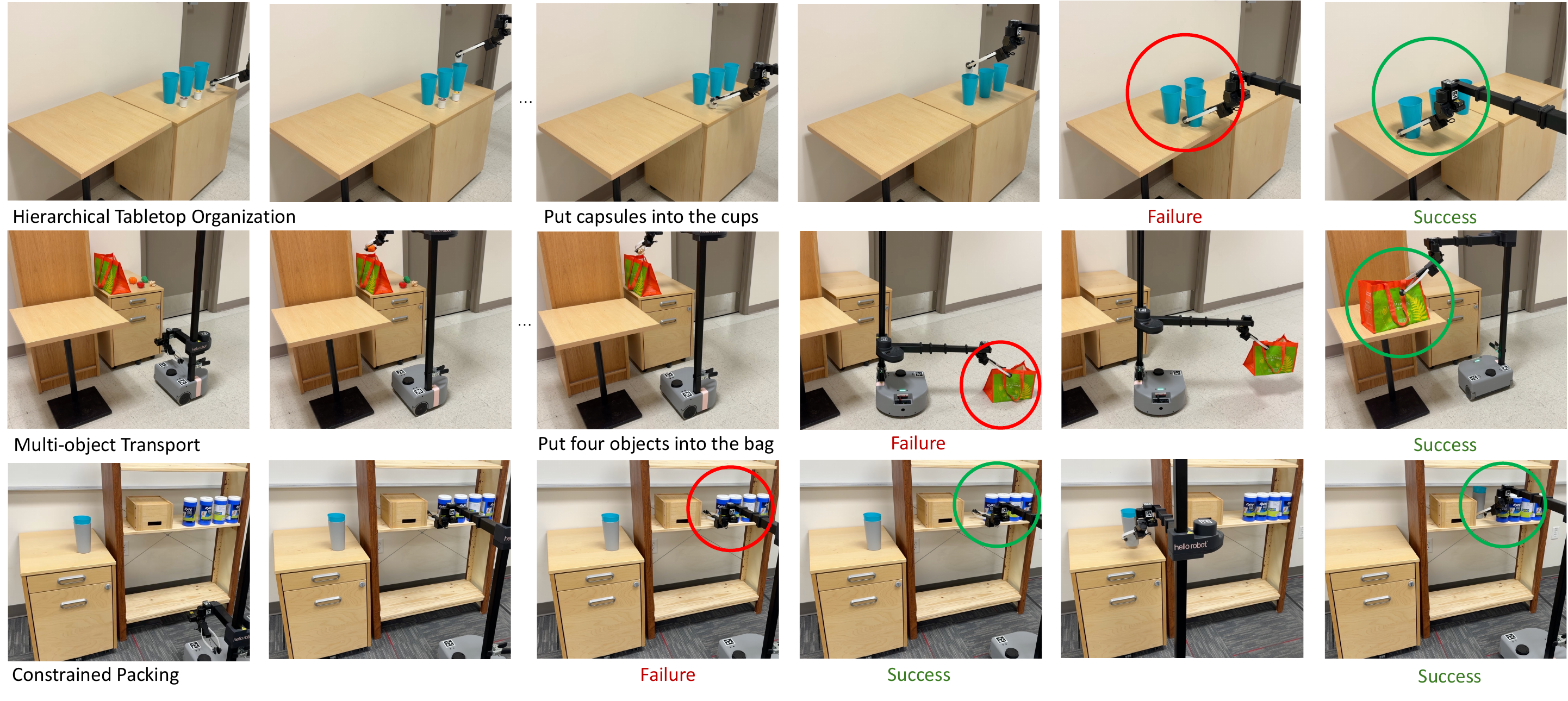}
    \caption{Rollouts of real-world evaluations and corresponding failure cases. A detailed explanation of this figure is provided in Sec.~\ref{sec:qualitative}.}
    \label{fig:real_world_comparison}
\end{figure}
\begin{figure}[th]
    \centering
    \includegraphics[width=0.99\columnwidth]{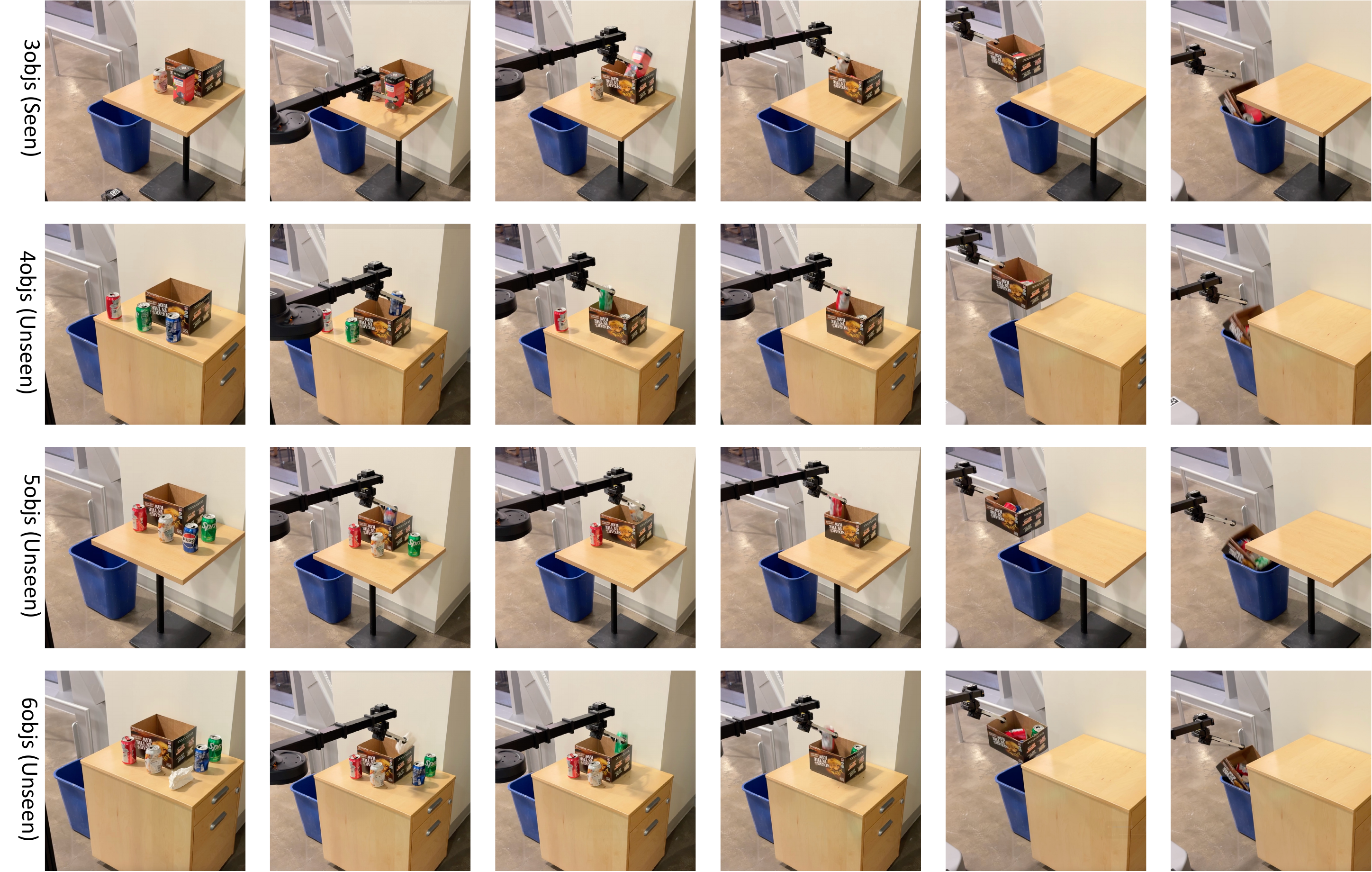}
    \caption{Real-world generalization visualizations. We show how \svgd{} generalizes to different numbers of objects~(3-6), different object shapes, and different tables.}
    \label{fig:real_generalization}
\end{figure}
We present qualitative results in Fig.~\ref{fig:real_world_comparison}. 
\textbf{Hierarchical Tabletop Organization} task~(First row): The robot is tasked with organizing the cups and capsules on another table while keeping them in a row. 
It first places several capsules into their corresponding cups. 
In the failure case, the robot fails to recognize the correlation between cups and capsules, resulting in the wrong organization. 
After learning from this failure, \svgd{} successfully completes this task by understanding that the capsules will move with their corresponding cups.
\textbf{Multi-object Transport} task~(Second row): The robot is tasked with packing groceries and placing them on the table. 
It places all four groceries inside a bag. 
In the failure case, the robot places the bag on the ground instead of the table, failing the task. 
After fine-tuning the model with a targeted dataset, \svgd{} moves the bag to the table. 
\textbf{Constrained Packing} task~(Third row): The robot is tasked with organizing a shelf by placing a stack of cups on a constrained shelf. 
In the failure case, the robot fails to make all the wipes in contact to clear enough space. 
After learning from the failure, \svgd{} first pushes the wipes aside in contact to create sufficient space for the cups, then places them on the shelf. 

Furthermore, we demonstrate how our approach generalizes to different numbers and shapes of objects, as well as different tables, in Fig.~\ref{fig:real_generalization}. 
Specifically, the model is fine-tuned only on failure cases with 3 objects but is able to generalize to scenarios involving 3-6 diverse objects on two tables.

\subsection{Detailed Experimental Tasks}\label{sec:task}
\textbf{Multi-object Transport} tasks the robot to transport multiple objects within a container using a single skill~(e.g., carrying multiple fruits in a grocery bag). 
To succeed, the robot has to understand that all objects inside the container move together when the container is moved. 
\textbf{Hierarchical Tabletop Organization} tasks the robot to organize a table by arranging objects into a hierarchical structure~(e.g., multiple objects in different cups). 
Success requires the robot to understand the relationships between these objects and how its skills impact future relations based on the hierarchical structure.  
\textbf{Constrained Packing} tasks the robot to organize objects in a constrained environment~(e.g., a bookshelf). 
Success involves using a non-prehensile push skill to create space and then packing the remaining objects onto the shelf. 
In this paper, we present quantitative results for the  \textbf{Multi-object Transport} and \textbf{Hierarchical Tabletop Organization} tasks, and qualitative results for the \textbf{Constrained Packing} task. 

\subsection{Efficiency Experiments}~\label{sec:efficiency}
\begin{wrapfigure}{r}{0.4\textwidth}
\vspace{-20pt}
   \includegraphics[width=0.95\linewidth,clip,trim=0mm 0mm 0mm 0mm]{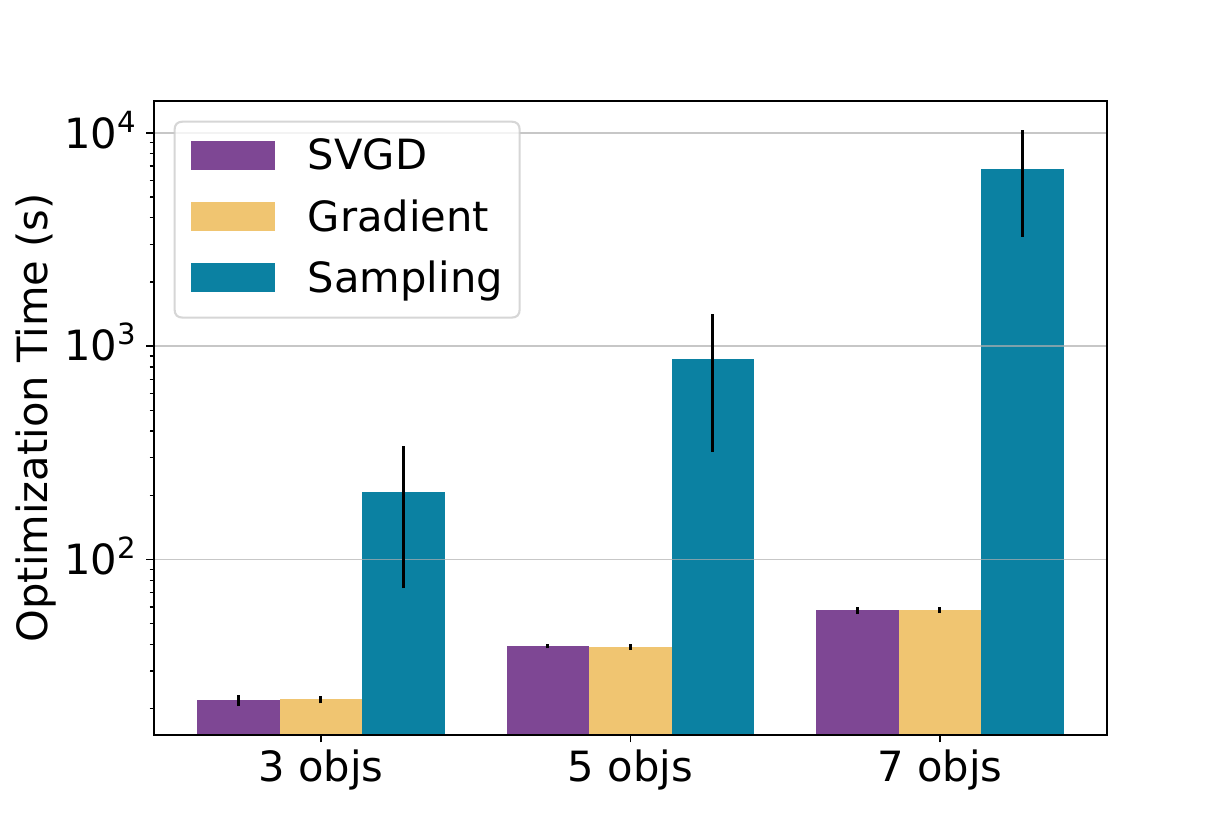} %
    \caption{Efficiency experiments show that \svgd{} is comparable to \gradient{} and more efficient than \sampling{}. 
    The optimization time is presented on a logarithmic scale. 
    \vspace{-10pt}
    }
    \label{fig:efficiency} 
    \vspace{-10pt}
\end{wrapfigure}
We compare \svgd{} with the two best-performing baselines, \gradient{} and \sampling, to assess optimization efficiency using the best-performing architecture~(\point{}). 
Error bars in the figure represent standard deviations across five different random seeds. 
As shown in Fig.~\ref{fig:efficiency}, \svgd{} is significantly more efficient than \sampling{}. 
This superior efficiency is attributed to the parallel computation capabilities of \stein{} on GPUs. 
\gradient{} achieves comparable efficiency to \svgd{}, but it still performs significantly worse in terms of fine-tuned model performance shown in Fig.~\ref{fig:real} and Table~\ref{table:simulation}. 

\subsection{Key Findings}\label{sec:key_findings}
\textbf{Importance of Learning from Failures:} Learning from failures is essential because an initial training dataset for a skill effect model cannot capture all possible transitions in the real world. When the robot encounters novel scenarios outside the training data distributions,
failures become inevitable. 
By learning from these failures, the robot can improve its performance more reliably and efficiently.

\textbf{Limitations of Replanning:} While \replan{} can recover from certain failures, its effectiveness is inherently limited. 
It relies on the learned dynamics model, and any inaccuracies in the model can significantly degrade its performance.  

\textbf{Effectiveness of \svgd{} with \stein{}:} Through large scale comparisons, we demonstrate that \svgd{} consistently outperforms the \sampling{} and \gradient{} baselines. This superior performance is attributed to \stein{}'s ability to approximate high-dimensional posterior distributions using multiple particles. 
In contrast, \sampling{} does not leverage gradient information, making it highly inefficient.
It also suffers from poor exploration with high-dimensional input. 
While \gradient{} uses gradient information, it suffers from mode collapse, leading to poor performance when the posterior distribution is multi-modal.

\textbf{Efficiency of \svgd{} with \stein{}:} Generating additional simulation datasets from observed failures requires solving a high-dimensional and multi-modal posterior distribution inference problem. 
\svgd{}, utilizing \stein{}, performs this task efficiently and achieves superior performance compared to \sampling{}. 
This efficiency is due to \stein{}'s ability to approximate complex posterior distributions in parallel, leveraging GPU computational power. 

Our experiments further reveal that as few as 20 targeted simulation data points are sufficient to fine-tune the skill effect model. 
This efficiency stems from the pre-trained model's ability to capture general representations that are transferable to related tasks. 
As a result, when the robot fails at a specific task, it can be efficiently fine-tuned on the new task to recover from failures. 

\textbf{Generalization of \svgd{}: } Through both simulation and real-world experiments, we demonstrate that our approach generalizes to varying numbers and shapes of objects, different environments~(e.g., tables), and viewpoints beyond those in the fine-tuning dataset. 
This demonstrates that our framework does not overfit to specific scenarios but instead captures object interaction in a generalizable manner. 
By combining the ability to continuously learn from failures and generalize to unseen scenarios, we believe our framework can adapt to diverse and complex real-world household environments, assisting in daily tasks.

\subsection{Ablation Study}\label{sec:ablation}
\begin{wrapfigure}{r}{0.4\textwidth}
\vspace{-10pt}
   \includegraphics[width=0.95\linewidth,clip,trim=0mm 0mm 0mm 0mm]{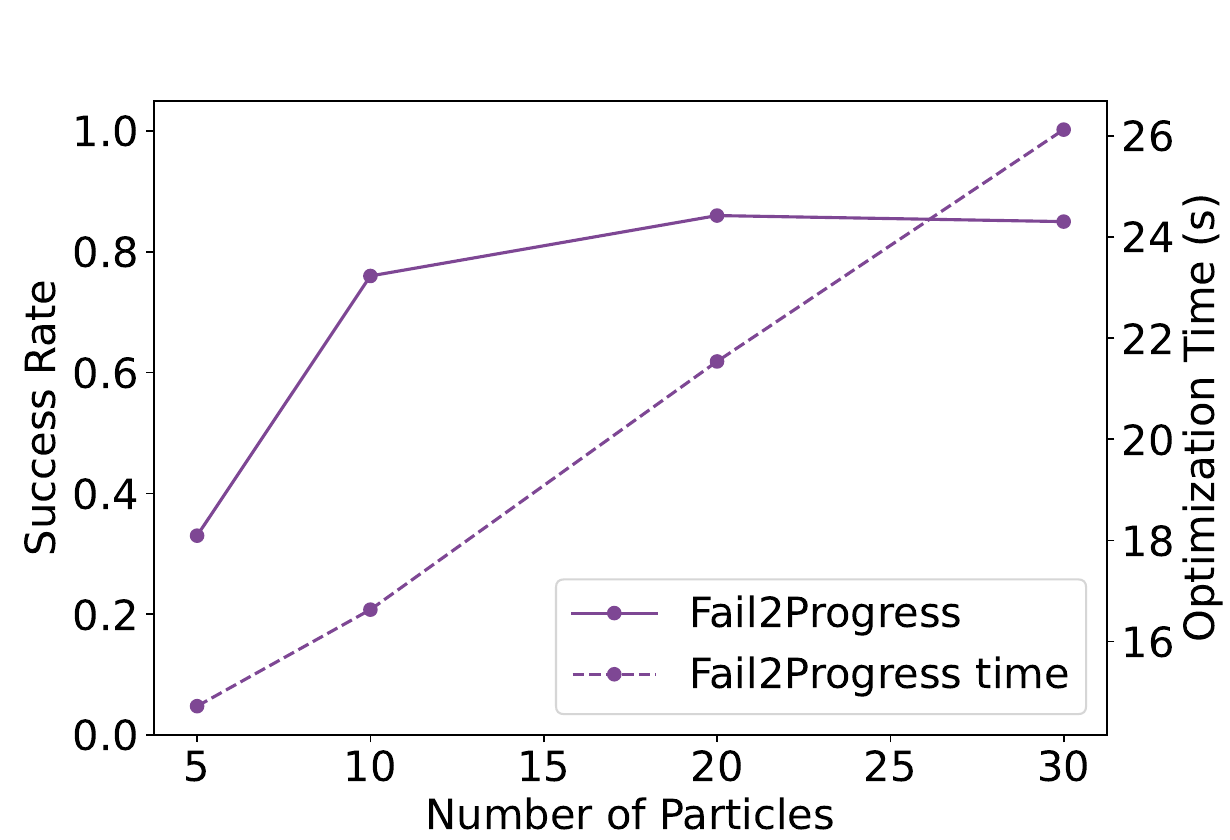} %
    \caption{An ablation study for different particles. 
    \vspace{-20pt}
    }
    \label{fig:particle_ablation} 
    \vspace{-10pt}
\end{wrapfigure}
To determine the best fine-tuning dataset size, measured by the number of particles, we conduct an ablation study presented in Fig.~\ref{fig:particle_ablation}. 
The study is performed on the \textbf{Hierarchical Tabletop Organization} task with three objects, using \point{} as the underlying architecture. 
Through the ablation study, we find that increasing the number of particles generally improves the execution success rate of \svgd{} but also increases the optimization time. 
Using 20 particles achieves the best balance between performance and efficiency. Therefore, we use 20 particles for our experiments.

\subsection{Detailed Simulation Results}\label{sec:detailed_simulation}
\begin{table}[h!]
    \centering
    \caption{\small
        Simulation experiments for the \textbf{Hierarchical Tabletop Organization} task across different numbers of objects. 
    The comparisons demonstrate that \svgd{} outperforms baselines by a large margin. 
    }
    \label{table:simulation_detailed}
    \adjustbox{max width=\textwidth}{
        \begin{tabular}{lcccccccc}
            \toprule
            \#Objs & Base &\orig{} & \sma{}  & \lar{}  & \replan{}  & \sampling{} & \gradient{} & \svgd{}\ \\
            \midrule
            3 & \point{}~\cite{huang2024points2plans} & 16\% & 18\% & 21\% & 28\% & 64\% & 52\% & \textbf{90\%} \\
            3 & Stow-GNN~\cite{chen2023predicting} & 13\% & 15\% & 19\% & 24\% & 61\% & 51\% & \textbf{83\%} \\
            3 & Binary-Pred~\cite{paxton-corl2021-semantic-placement} & 11\% & 13\% & 15\% & 23\% & 51\% & 47\% & \textbf{78\%}  \\ \bottomrule 
            5 & \point{}~\cite{huang2024points2plans} & 10\% & 13\% & 15\% & 26\% & 54\% & 45\% & \textbf{87\%} \\
            5 & Stow-GNN~\cite{chen2023predicting} & 9\% & 10\% & 14\% & 21\% & 52\% & 43\% & \textbf{81\%} \\
            5 & Binary-Pred~\cite{paxton-corl2021-semantic-placement} & 8\% & 8\% & 13\% & 19\% & 43\% & 37\% & \textbf{73\%} \\ \bottomrule
            7 & \point{}~\cite{huang2024points2plans} & 8\% & 9\% & 12\% & 18\% & 42\% & 39\% & \textbf{82\%} \\
            7 & Stow-GNN~\cite{chen2023predicting} & 8\% & 7\% & 11\% & 17\% & 41\% & 37\% & \textbf{76\%} \\
            7 & Binary-Pred~\cite{paxton-corl2021-semantic-placement} & 5\% & 7\% & 9\% & 14\% & 30\% & 29\% & \textbf{64\%} \\ \bottomrule
            Average & \point{}~\cite{huang2024points2plans} & 11\% & 13\% & 16\% & 24\% & 53\% & 45\% & \textbf{86\%} \\
            Average & Stow-GNN~\cite{chen2023predicting} & 10\% & 11\% & 15\% & 21\% & 51\% & 44\% & \textbf{80\%} \\
            Average & Binary-Pred~\cite{paxton-corl2021-semantic-placement} & 8\% & 9\% & 12\% & 19\% & 41\% & 38\% & \textbf{72\%} \\ \bottomrule
        \end{tabular}
    }
\end{table}
We provide the detailed simulation results in Table~\ref{table:simulation_detailed}. Through the comparison, we find that \svgd{} outperforms all baselines with different numbers of objects using different base architectures. 

\subsection{Detailed Sim2Real Gap}\label{sec:sim2real}
We provide a detailed analysis of the Sim2Real gap for our work. The Sim2Real gap can be caused by the following reasons: 
(1): Perception gap: This includes differences in rendering and visualization between simulation and the real world~(e.g., real-world perception is less accurate and noisier). 
In simulation, ground-truth object segmentation are readily available, whereas in the real world, obtaining accurate information is challenging, especially for partial views and cluttered scenes.  
(2): Controller mismatch: This arises due to differences in robot control between simulation and the real world. Real robots would have latency, compliance, and joint limit constraints, which are often not fully modeled in simulation. 
(3): Object geometry gap: Real-world objects vary in material, shape, and appearance, and often differ from their simulation twin. This discrepancy is particularly significant for deformable objects. 
(4): Physical modeling: The physical modeling in simulation can be inaccurate, which contributes to unrealistic physical interactions.

\subsection{Extra Related Work}\label{sec:extra_related}
Solving long-horizon manipulation tasks remains a significant challenge in the community. 
Traditionally, task and motion planning~\cite{garrett2021integrated} addresses this problem by separating high-level symbolic reasoning from low-level geometric reasoning. 
However, TAMP methods typically rely on explicit 3D object models~\cite{liang-icra2022, garrett2020pddlstream, garrett2017sample, garrett2021integrated, kim2019learning} and symbolic operators with predefined effects~\cite{curtis2022long, garrett2020pddlstream, garrett2017sample, garrett2021integrated, kim2019learning, driess2020deep}. 
Alternatively, recent works propose to sequence learned skills to handle geometrically complex tasks, but these approaches are also limited to hand-crafted states~\cite{stap, lin2023text2motion, gsc, cheng2023league}. 
A more recent study~\cite{huang2024points2plans} introduces a method to learn the effects of skills directly from partial-view point clouds, enabling robots to reason about real-world scenarios involving hard-to-predefine object interactions. 
However, none of the existing skill effect models~\cite{stap, lin2023text2motion, gsc, cheng2023league, huang2024points2plans} reason about or learn from failures after deployment. 
Our work is the first to leverage the failure cases to improve skill effect models, thereby minimizing future failures. 

\subsection{Relations Definition}\label{sec:relations}
We use both unary relations and binary relations in this paper. 

We consider the following unary relations: (1) \texttt{Movability}: indicates whether a specific segment is movable~(e.g., a table is not \texttt{movable}). 
(2) Drawer identification: specifies whether a segment is a drawer. 
(3): Drawer state: determines whether a drawer is \texttt{open} or \texttt{closed}. 

We define the following binary relations: 
(1) Spatial relations: includes six spatial relationships-\texttt{left, right, front, behind, above, below}- defined following~\cite{paxton-corl2021-semantic-placement}. 
(2) \texttt{In-contact}: identifies whether two objects are in contact. 
Ground-truth labels for this relation are obtained directly from the IsaacGym~\cite{isaacgym} simulator. 
(3) \texttt{Boundary}: Specifies whether an object is on the boundary of a supporting surface~(e.g., a table or shelf). 
We define \texttt{boundary}(A, B) as true if 
object A is above object B, the distance between A and the nearest boundary of B is less than a threshold $\epsilon_{boundary}$, and the dimensions of B exceed $\epsilon_{bottom}$. 
In this paper, $\epsilon_{boundary}$ is set to 0.1m, and $\epsilon_{bottom}$ is set to 0.2m. 
(4) \texttt{Inside}: indicates whether an object is inside another~(e.g., a container). We define \texttt{Inside}(A, B) as true if the bounding box of A is completely contained within the bounding box of B.

\subsection{Skills Definition}\label{sec:skills}
We use the following skills in this paper. 

\textbf{Pick-and-place}: This skill enables the robot to grasp an object and place it in a specific pose. 
If the grasp pose or placement pose is outside the robot's current reachable space, the robot will first move its base to make it reachable before executing the arm motions. 
The continuous parameter encodes the difference between the placement pose and the grasp pose. 

\textbf{Push}: This skill allows the robot to push multiple objects. 
The robot first moves to a pre-push pose and then moves its end-effector along the push direction for a specific distance.
The continuous parameter encodes both the push direction and push distance. 

\textbf{Open/Close Drawer}: This skill enables the robot to open or close a drawer. 
The corresponding continuous parameters encode the distance and direction of the motion. 

Notably, if failures occur with a newly introduced skill, a new skill effect model can be trained to handle that skill effectively. Due to the composability of the skill effect model, the planning can incorporate all the skills. 

\subsection{Details of Skill Effect Models}\label{sec:SEM}

\subsubsection{Introduction}
Given an observation, $O_t$, at time $t$, represented as segmented point clouds, the skill effect model encodes \(O_t\) to a latent state $X_t$ using $\encoder{}$. 
Using a decoder, $\decoder{}$, the latent state can be decoded to either geometric states like object poses or symbolic states such as inter-object relations, $\mathcal{R}$. 
Furthermore, the latent state, $X_t$, could also be propagated by a skill $\phi_t(a_t)$ with a dynamics model $\dynamics{}$ to predict the latent state $X_{t+1}$ at the next time step. 
The predicted latent state $X_{t+1}$ could also be decoded to predicted object poses or relations. 
To simplify, in this paper, we use $\gamma(\cdot)$ to represent the skill effect model composing the different components $\encoder{}$, $\decoder{}$, and $\dynamics{}$ as $\gamma(O_t, \phi_t, a_t) = \decoder{}(\dynamics{}(\encoder{}(O_t), \phi_t(a_t)))$, that outputs the probabilities of different relations in $\mathcal{R}$, with $\Gamma(O_0, \phi_{1:H}, a_{1:H}) = \gamma_H \circ \gamma_{H-1} \circ  ...  \circ \gamma_1 $, representing a composition of skill effect for a skill sequence $\phi_{1:H}(a_{1:H})$. 

\subsubsection{Implementation Details}
The input of a skill effect model~(e.g., \point{}) is a segmented point cloud at timestep t, denoted as $O_t = \{O_t^0, ..., O_t^{n}\}$, where n represents the number of segments.  

\textbf{Encoder:}
We utilize PointConv~\cite{wu2019pointconv} as the $\encoder$. 
The employed PointConv architecture consists of three set abstraction layers, each processing input point data and corresponding positional data to produce sampled positional data and feature data as output. 
Both the input and output positional data have three channels. 
The first abstract layer samples 128 points, with 8 neighbors per point determined using a bandwidth of 0.1. 
It employs an MLP with 6 input channels~(3 for positions and 3 for features), 32 output channels, and a kernel size of 1. 
The second layer reduces the sample size to 16 points with 16 neighbors per point and uses a bandwidth of 0.2. 
This layer's MLP takes 35 input channels~(3 for positions and 32 for features), and outputs 64 channels with a kernel size of 1. 
The third layer is a "group all" layer that generates 128-dimensional features per segment,  using a bandwidth of 0.4, 
Its MLP has 67 input channels~(3 for positions and 64 for features) and 128 output channels, with a kernel size of 1. 

Specifically, the encoder~($\encoder$) processes each segment to generate a corresponding point cloud feature, represented as $P_t^i = \encoder(O_t^i)$, where each point cloud feature has 128 dimensions.  
Additionally, we use positional encoding in PyTorch~\cite{NEURIPS2019_9015} to assign a unique identifier to each object, represented as $ID_{i}$, which also has 128 dimensions. 
For each object, we concatenate the point cloud feature and positional encoding to form $X_t^i = P_t^i \oplus ID_{i}$, resulting in a feature vector with 256 dimensions. 
Consequently, the latent state is represented in an object-centric form as $X_t = \{X_t^0, ..., X_t^{n}\}$, where each object's latent state contains 256 features. 

\textbf{Dynamics:}
The dynamics model~($\dynamics$) takes as input the latent state $X_t$ along with the corresponding skill and continuous parameter $\phi_1(a_1)$. 
Since $\phi_1$ encodes discrete parameter identifying the object to manipulate, we use positional encoding to represent the manipulated object ID as $ID_{i}$, which has 128 features. 
For the continuous parameter $a_1$, we use a simple MLP $MLP_{para}$ to encode a latent continuous parameter with 128 features. 
The $MLP_{para}$ consists of two layers, each with 128 neurons, using ReLU as the activation function. 
As a result, each skill is represented as a latent state $AL_1 = MLP_{para}(a_1) \oplus ID_{i}$, where $AL_1$ has 256 features. 

Once the latent skill $AL_1$ is obtained, we use a transformer-based dynamics model, $\dynamics$.
The transformer comprises 2 sub-encoder layers, 2 attention heads in the multi-head attention mechanism, and a dimensionality of 256 for the input and output. 
Given the latent state $X_t$ and the corresponding latent skill $AL_1$, $\dynamics$ outputs the change in each latent state, represented as $\delta X_t$. 
The predicted new latent state is then computed as $X_{t+1} = X_t + \delta X_t$. 
For long-horizon planning, the dynamics model can be applied recurrently as $X_{t+H} = \dynamics(X_t, AL_{1:H})$. 

\textbf{Decoder:} 
The decoder~($\decoder$) consists of three distinct modules: a position decoder $\decoder_{p}$, a unary relation decoder $\decoder_{u}$, and a binary relation decoder $\decoder_{b}$. 
All decoders take the latent state~($X_t$) as input. 

Position Decoder~($\decoder_{p}$): The position decoder processes the predicted changes in each latent state~($\delta X_t$) and outputs the predicted changes in object positions~($\delta p_t$). 
$\decoder_{p}$ is a three-layer network, with each layer containing 64 neurons and ReLU as the activation function. 

Unary relation decoder~($\decoder_{u}$): The unary relation decoder takes the absolute latent state~($X_t$) as input and outputs unary object relations. 
$\decoder_{u}$ consists of two layers, each with 64 neurons, and uses Softmax as the activation function because unary relations are binary variables. 

Binary relation decoder~($\decoder_{b}$): The binary relation decoder takes pairwise latent states~($(X_t^i, X_t^j)$) as input and outputs pairwise object relations, defined as $\decoder_{b}(X_t^i, X_t^j)$. 
$\decoder_{b}$ is a three-layer network, with each layer containing 64 neurons. 
Like $\decoder_{u}$, $\decoder_{b}$ uses Softmax as the activation function since binary relations are also binary variables. 

\textbf{Training Details:}
We collect ground-truth data from the simulation at the current step, including point cloud observations~($O_t$), relations~($\mathcal{R}_t$), and position~($p_t$). 
Ground-truth data at the next step is collected after executing a robot skill, which includes point cloud observations~($O_{t+1}$), relations~($\mathcal{R}_{t+1}$), and position~($p_{t+1}$). 
To train the skill effect model using the simulation dataset~($\dataset$), we employ several loss functions: 

Current step detection loss: Using the current step point cloud observation~($O_t$), the model~($\rd$) predicts current step relations~($\hat{\mathcal{R}}_t$). The current step detection loss is calculated as $L_{detection} = CE(\hat{\mathcal{R}}_t, R_t)$, where $CE$ denotes the cross-entropy loss. 

Latent space regularization loss: 
The mode encodes the observations ($O_t$, $O_{t+1}$) into the current step latent state~($X_t$) and the next step latent state~($X_{t+1}$). 
Using a skill, $\rd$ predicts the next time step latent state~($X^{\prime}_{t+1}$), where $X^{\prime}_{t+1}$ is derived from $O_t$ and the skill while $X_{t+1}$ derives from $O_{t+1}$. 
The regularization loss, calculated as the L2 norm, is $L_{regulization} = ||X_{t+1} - X^{\prime}_{t+1}||^2_2$.

Position loss:
Based on $O_t$ and the applied skill, the model predicts the change in object positions~($\delta p_t$).
The position loss compares the predicted position changes with the ground-truth position changes: 
$L_{pos} = b \cdot \sqrt{a\cdot ||\delta p_t - (p_{t+1} - p_t)||}$. 
Here, $a = 12$ and $b = 5$ are used to balance other loss terms, as defined in~\cite{huang2024points2plans}. 

Prediction loss: 
To minimize the difference between predicted relations~($R^{\prime}_{t+1} = \decoder_{b}(X^{\prime}_{t+1})$) and ground-truth relations~($R_{t+1}$) at the next time step, we compute the prediction loss as: $L_{prediction} = CE(R^{\prime}_{t+1}, R_{t+1})$. 

The total loss is the sum of all four terms: 
$L = L_{detection} + L_{regulization} + L_{pos} + L_{prediction}$.
We train and fine-tune the skill effect model using the Adam optimizer with a learning rate of $1 \times 10^{-4}$. 
In this paper, we use 10 epochs for the pre-training and 200 epochs for the fine-tuning. 

\textbf{Planning Details:}
To achieve the goal relations~($\mathcal{G}$), we employ a shooting-based approach to sample the continuous parameters~($a_{1:H}$) given the initial observation~($O_1$) and the plan skeleton~($\phi_{1:H}$). 
To maximize the likelihood of achieving the goal relations, we sample a set of continuous parameters $\{a^{j}_{1:H}\}_{j=1}^{Ka}$ from the robot's workspace. 
Each continuous parameter sequence $a^{j}_{1:H}$ is rolled out, and we select the sequence that maximizes the probability of satisfying $\mathcal{G}$.

\subsection{Real-to-sim details}\label{sec:real2sim_details}
For the real-to-sim process, \stein{} generates both simulation states and robot skills. 
The simulation states specify the pose of each object. 

To create a simulation scene, we assume a set of object shape priors, including cuboids, open boxes, shelves, drawers, and tables. 
Based on the semantics of each segment in the observation, our method selects the appropriate object shape prior. 
The bounding box of each segment determines the dimension of the corresponding object in the simulation. 
By combining these dimensions with the object poses, we can construct the simulation scene. 

For the robot skills, we directly execute the parameterized skills within the simulation, starting from the initial scene. During the process, we could record point clouds and object relations both before and after manipulation. Combined with the executed robot skills, we can generate a fine-tuning dataset to refine the skill effect model. 

Note that we select this bounding-box approximation for the real-to-sim approach due to efficiency considerations. However, \svgd{} can compliment other real-to-sim approaches~\cite{chen2024urdformer, mandi2024real2code, torne2024reconciling}.

\subsection{Stein Update Details}\label{sec:stein_details}
\subsubsection{Generating State Samples}
First, we aim to solve for the simulation state set $S^{+}$. 
Here we want to find samples $q(S) = \{s_i^{+}\}_{i=1}^M$ that approximate the posterior distribution $P(r^{\fail}\mid O^{+} = \xi(S)O^{\fail})P(S)$, where $P(S)$ is a uniform prior over all feasible simulation states. 
The posterior distribution ensures that the transformed point clouds match the relations in the failure case $r^{\fail}$. 
This defines the following variational inference problem:  
\begin{align}\label{eq:state-KL}
    \argmin_{q(S)} \,
    & \; \dkl{q(S)}{\rd(r^{\fail}\mid O^{+} = \xi(S)O^{\fail})P(S)}
\end{align}

For each state particle $s^{+}_i$, the Stein update term is:
\begin{equation}\label{eq:Stein-2}
 \Phi(s_i^+) = \frac{1}{M}\sum_{j=1}^{M} [k(s^{+}_j, s_i^{+})  \nabla_{s^{+}_{j}}\ln P(\mathcal{R}^{\fail}\mid \xi(s^{+}_{j})O^{\fail}) +             \nabla_{s^{+}_{j}}k(s^{+}_{j}, s_i^{+})]
\end{equation}
where $k(s^{+}_{j}, s_i^{+})$ is a kernel function that defines the similarity between different particles. The first term in Eq.~\ref{eq:Stein-2} represents an attractive force that pushes the particles to move in a direction based on the gradient while the second term is a repulsive term that prevents the particles from collapsing. 
This update can generate object states that match the failure case while ensuring diversity over object states.  

\subsubsection{Generating Action Samples} 
Given our state samples generated by using Stein variational inference to approximate the distribution in Eq.~\ref{eq:approx-constraints}, we can now turn our attention to solving for the action set \(A^{+}\). To formulate this problem we make use of the generalized Bayesian inference framework outlined above. Here we define the loss function, \(\mathcal{L}\), to be the entropy loss defined in Eq.~\ref{eq:approx-objective} and let \(\beta=1\). Note that the variational distribution \(q(A) = \{(s^{+}_i, a^{+}_i)\}_{i=1}^M\), however we keep the values of \(s_i^{+}\) fixed and search only over actions. This defines the following variational inference problem: 
\begin{equation}\label{eq:action-KL}
 \argmin_{q(A)} \meandist{\prod_{r \in \mathcal{R}^{\fail}} -H(\rd(r\mid \xi(s^{+})O^{\fail}, \phi^{\fail}, a^{+}, D))}{s^+, a^+ \in q(A)} + \quad \dkl{A^{+}}{P(A)}
\end{equation}
where $P(A)$ is uniform prior over actions. 

The Stein update term for the action particles $a^+_i$ is:
\begin{equation}\label{eq:Stein-action_2}
\Phi(a^+_i) = \frac{1}{M}\sum_{j=1}^{M} [k(a^{+}_j, a^{+}_i) \cdot \nabla_{a^{+}_{j}}\ln H(\rd(\mathcal{R}^{\fail}\mid \xi(s^{+}_{j})O^{\fail}, \phi^{\fail}, a^{+}_{j})) + \nabla_{a^{+}_{j}}k(a^{+}_{j}, a^{+}_i)]
\end{equation}

\subsubsection{Implementation Details of \stein{}}
We use RBF kernels for \stein{} and follow 
previous works~\cite{pavlasek2023ready, garreau2017large} by applying the median heuristics to determine the kernel bandwidth. Additionally, the step size is optimized using the Adam optimizer~\cite{kingma2014adam}. 

\subsection{Detailed Generalization Experiments}\label{sec:generalization}
\begin{figure}[th]
    \centering
    \includegraphics[width=\columnwidth]{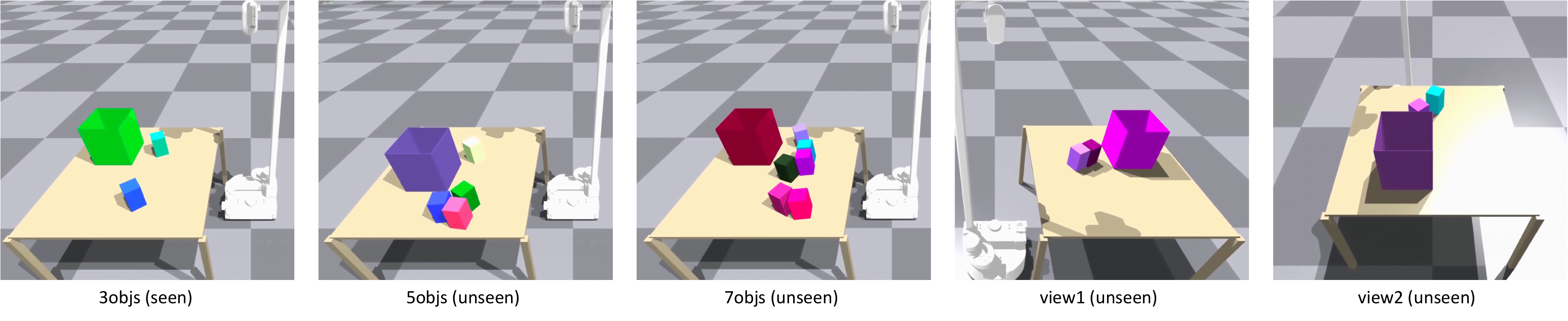}
    \caption{Visualizations of simulation generalization scenarios. \svgd{}, fine-tuned on a dataset with 3 objects, successfully generalizes to scenes with 5 and 7 objects. Additionally, \svgd{} demonstrates generalization to two unseen viewpoints.}\label{fig:sim_visualization_generalization}
\end{figure}

\begin{wraptable}{rt}{0.5\textwidth}
    \centering
    \caption{\small
        Generalization results for the \textbf{Multi-object Transport} task. We show the generalization capability of \svgd{} with respect to different numbers of objects and different viewpoints~(\btext{5objs, 7objs, view1, and view2} are unseen in the training dataset). Evaluations on unseen objects and unseen viewpoints show that \svgd{} performs well and outperforms the best-performing baselines~(\sampling{} and \gradient{}).
        \vspace{-10pt}
    }
    \label{table:generalization}
    \adjustbox{max width=\textwidth}{
        \begin{tabular}{lccccc}
            \toprule
            Generalization Scenarios & 3objs $\uparrow$ & \bcl{5objs} $\uparrow$ & \bcl{7objs} $\uparrow$ & \bcl{view1} $\uparrow$ & \bcl{view2} $\uparrow$\\ \\
            \midrule
            \svgd{}  &  \textbf{87\%} & \textbf{81\%} & \textbf{71\%} & \textbf{83\%} & \textbf{85\%}\\
             \gradient{}  &  51\% & 40\% & 18\% & 42\% & 44\%\\
             \sampling{}  &  62\% & 45\% & 23\% & 51\% & 47\%\\
            \bottomrule
        \end{tabular}
    }
\end{wraptable}
\textbf{Generalization Evaluation: }
We assess the generalization capability of \svgd{} compared to the \gradient{} and \sampling{} baselines in the \textbf{Multi-object Transport} task. 
First, we evaluate generalization to an unseen number of objects, as shown in Table~\ref{table:generalization}. 
The model is fine-tuned only on scenarios with 3 objects and tested on unseen scenarios with 5 and 7 objects. 
While all approaches experience some performance degradation, \svgd{} maintains strong performance, even in scenarios with 7 objects. 
In contrast, \gradient{} and \sampling{} perform poorly, particularly in the 7-object scenarios. 
Next, we assess generalization to unseen viewpoints, also shown in Table~\ref{table:generalization}. 
\svgd{} demonstrates robust performance across two unseen viewpoints and consistently outperforms \gradient{} and \sampling{} baselines.  
For both evaluations, we perform 100 trials per approach for each evaluation metric. Visualizations of these generalization scenarios are provided in Fig.~\ref{fig:sim_visualization_generalization}. 

\subsection{Experimental Details} 
We now provide further details regarding the datasets. 
We first report the failure relations satisfaction score, which measures the percentage of relations in each dataset that match failure cases. 
These scores for Fail2Progress, Original, Small, Large, Sampling, and Gradient are \textbf{96.2\%}, 82.4\%, 82.9\%, 82.8\%, 89.7\%, and 90.4\%, respectively. 
Second, we report the standard deviation of the action parameters to evaluate the diversity of continuous action values. 
The corresponding standard deviations of the action parameters are \textbf{0.673m}, 0.415m, 0.421m, 0.424m, 0.523m, and 0.347m.
These details support the findings presented in the main paper. The Original, Small, and Large baselines perform poorly because their training datasets do not capture out-of-distribution failures, resulting in lower failure satisfaction and action diversity scores. 
In contrast, Fail2Progress achieves the highest failure relation satisfaction and exhibits the most diverse actions, outperforming all baselines, including Gradient and Sampling. We attribute this superior performance to the ability of SVI to approximate high-dimensional, multi-modal posterior distributions. 
Note that we use rejection sampling as the sampling baseline, and it is significantly slower than Fail2Progress (Appx. A.3). 

\subsection{Additional Baseline for Domain Randomization}
Domain randomization is a technique used to address failures arising from the Sim2Real gap. If one considers randomized objects, poses, and environments as domain randomization, then all of our \sma{} and \lar{} baselines can be considered domain randomization baselines. 

\begin{wraptable}{rt}{0.5\textwidth}
    \centering
    \caption{\small
        Comparison against a domain randomization baseline using \point{} architecture.
        \vspace{-5pt}
    }
    \label{table:domain_r}
    \adjustbox{max width=\textwidth}{
        \begin{tabular}{lccccc}
            \toprule
             & 3objs $\uparrow$ & 5objs $\uparrow$ & 7objs $\uparrow$  \\
            \midrule
            \svgd{}  &  \textbf{90\%} & \textbf{87\%} & \textbf{82\%} \\
             Domain randomization  &  59\% & 49\% & 36\% \\
            \bottomrule
        \end{tabular}
    }
\end{wraptable}
Furthermore, we conduct experiments where actions are held constant while object states and environments are randomized to generate failure-targeted training data using~\point{} architecture. 
The average success rate for this domain randomization is 48\%, which is significantly lower than our proposed approach, Fail2Progress, which achieves an average success rate of 86\%. The comparison details are shown in Table.~\ref{table:domain_r}. We conduct 100 trials for each approach and each object count.

\subsection{Hardware Information}
All the skill effect models are trained and fine-tuned on a standard workstation with an NVIDIA GeForce RTX 3090 Ti GPU. 
All the real-world experiments are conducted with a Stretch-re2 from Hello-Robot.

\end{document}